\documentclass[runningheads]{llncs}
\usepackage[T1]{fontenc}
\usepackage{graphicx}
\usepackage{algorithm}
\usepackage{algorithmic}

\begin{document}
\title{On Explaining Proxy Discrimination\\ and Unfairness in Individual Decisions \\ Made by~AI~Systems}
\titlerunning{Formal XAI for Proxy Discrimination and Unfairness}
\author{Belona Sonna\inst{1}\orcidID{0000-0001-5682-746X} \and
Alban Grastien\inst{2}\orcidID{0000-0001-8466-8777}}
\authorrunning{B. Sonna and A. Grastien}
\institute{The Australian National University, Acton, Canberra, Australia\and
Université Paris-Saclay, CEA, List, F-91120, Palaiseau, France }

\maketitle              % typeset the header of the contribution
\begin{abstract}
Artificial intelligence (AI) systems in high-stakes domains raise concerns about proxy discrimination, unfairness and explainability. Existing audits often fail to reveal why unfairness arises, particularly when rooted in structural bias. We propose a novel framework using \emph{formal abductive explanations} to explain proxy discrimination in individual AI decisions. Leveraging \emph{background knowledge}, our method identifies which features act as unjustified proxies for protected attributes, revealing hidden structural biases. Central to our approach is the concept of \emph{aptitude}, a task-relevant property independent of group membership, with a \emph{mapping function} aligning individuals of equivalent aptitude across groups to assess fairness substantively. As a proof of concept, we showcase the framework with examples taken from the German credit dataset, demonstrating its ability to be used in real world case. 
\end{abstract}

\section{Introduction}

Artificial intelligence (AI) is increasingly embedded in decision-making across finance, healthcare, and criminal justice~\cite{Ferrer2021}. However, as these systems enter high-stakes domains, persistent concerns arise: \textbf{proxy discrimination}, where neutral features encode protected attributes~\cite{lindholm2024fair}; \textbf{unfairness}, where individuals are treated unequally on such grounds~\cite{lindholm2024fair}; and \textbf{lack of transparency}, as complex models resist interpretation~\cite{VANDERWAA2021103404}. 
These challenges are now central to both public debate and technical research~\cite{barocas2016,mittelstadt2016ethics}.

Among these, proxy discrimination has received particular attention. 
Neutral inputs may act as stand-ins for sensitive attributes, producing discriminatory outcomes. 
Research has approached this problem from multiple directions: \emph{formal tools} for detecting proxy use, often rooted in causal reasoning~\cite{weerts2024unlawful}; \emph{proxy groups} as fairness interventions when protected features are unavailable~\cite{gupta2018proxy,lindholm2024fair}; and \emph{conceptual analyses} clarifying the normative and epistemic status of proxies~\cite{johnson2025hardproxy,tschantz2022proxy}. 
Despite this diversity, consensus holds that proxy use introduces problematic predictive dependencies. 
What remains unsettled is a single formal account of when and why such use becomes discriminatory.

Closely related is the broader agenda of \textit{fairness auditing}, which seeks to evaluate and mitigate algorithmic bias. Approaches include \emph{dataset audits} that assess representativeness, and \emph{model audits} that apply fairness metrics such as demographic parity, equalized odds, or calibration across groups. Causal and counterfactual audits extend these methods by uncovering unfair pathways~\cite{gebru2018datasheets,tschantz2022proxy,weerts2024unlawful,liu2025fairsense}. Despite methodological progress, consensus holds that fairness auditing requires transparency, contextual evaluation, and explanatory insight, not just statistical checks.

Transparency, however, remains a major challenge. Clear explanations improve trust, reveal hidden dependencies, and support accountability~\cite{wachter2017counterfactual}. Although regulatory frameworks such as the EU’s GDPR established a ``right to explanation''~\cite{EU2016}, the black-box nature of many models continues to obstruct oversight~\cite{shih2018symbolic,Emmert_Streib_2020}. 
This motivates the search for approaches that not only assess whether a system is fair but also explain \emph{why} unfairness arises, especially when rooted in structural or proxy-based discrimination. We propose a formal framework that employs \emph{abductive explanations} to diagnose proxy discrimination and unfairness in AI-based decisions.

Within formal explainable AI (XAI), abductive explanations complement contrastive or counterfactual explanations~\cite{formalXAI2021}. Counterfactuals excel at answering \emph{what-if} scenarios and offering recourse, while abductive explanations provide logical proofs explaining why a decision occurred~\cite{harman1965inference,keith2017abductive}. They answer the question: \emph{Why does the system produce outcome $d$ for individual $x$}? By logically computing plausible hypotheses that explain observed outcomes~\cite{harman1965inference}, abductive explanations become a powerful tool for diagnosing discrimination by identifying causes embedded in decision outcomes.

Recent studies highlight the promise of abductive reasoning for debugging and bias detection~\cite{lambers2020abductive,shih2019symbolic}. A decision is biased if all sufficient explanations for it include a protected attribute~\cite{Adnan2020Reasons}. This instance-level perspective enables the detection of proxy discrimination beyond aggregate statistical checks. Building on this foundation, our contributions are threefold: we formally define proxy variables within abductive explanations (Section~\ref{subsection::proxy-variable}); we redefine bias in AI decisions by incorporating proxy effects and real-world constraints (Section~\ref{subsection::proxy-discrimination}); and we introduce a fairness assessment framework based on abductive explanations that systematically captures subgroup inequalities (Section~\ref{section::unfairness}).

\section{Notation and Preliminaries}
\subsection{Notation}
We use the following notation throughout:
\begin{itemize}
  \item $V = \{v_1, \dots, v_n\}$ denotes a set of $n$ Boolean variables (features).
  \item An \emph{individual} $x = [x_1, \dots, x_n]$ is represented as a vector of Boolean values corresponding to the features in $V$, also interpreted as a set of literals. 
  \item $X$ denotes the space of all possible individuals.
  \item $D$ is the domain of the decision function, with $d \in D$.
  \item $\Delta: X \rightarrow D$ is the \emph{decision function} computed by the model.
  \item $XP$ is a \emph{property} of an individual $x$, defined as a subset (i.e., conjunction) of literals from $x$, such that $XP \subseteq x$. We denote by $vars(XP)$ the set of features that appear in $XP$.
  \item $p \in V$ denotes the \emph{protected} variable.
  \item $X_{(V')}^x$ denotes the set of individuals that are identical to $x$ except (potentially) on the subset of features $V' \subseteq V$.
\end{itemize}
\subsection{Running Example}
To ensure clarity, we introduce a running example based on the German Credit Dataset. The decision process is modeled by a binary classifier ($\Delta$) that predicts whether an individual is a good or bad credit risk ($R$) based on the following features: Age ($A$), Gender ($G$), Job ($J$), Housing ($H$), Savings ($S$), Account Balance ($B$), Credit Amount ($C$), Duration of Last Credit ($D$), Credit Purpose ($P$), and Marital Status ($M$). The protected feature is \emph{gender} ($G$). Additional details on the dataset and decision process can be found in Appendix~\ref{app:decision_process}.

\subsection{Abductive Explanation and Computation}
Abductive explanations have been extensively studied in the literature~\cite{Marques-Silva_Ignatiev_2022,shih2018symbolic,Adnan2020Reasons}. An explanation is defined as a minimal set of features of an individual that is sufficient to guarantee the same decision. That is, any individual sharing this set of features will receive the same output from the decision function.

\begin{definition}[Abductive Explanation]\label{def::AXP}
    Let $x$ be an individual with decision $d = \Delta(x)$. A subset $XP \subseteq x$ is an \emph{abductive explanation} for decision $d$ if:
\begin{displaymath}
    \forall x' \in X.\quad XP \subseteq x' \Rightarrow \Delta(x') = d.
\end{displaymath}
\end{definition}
%begin belona
In this work, we are interested in \emph{subset-minimal} explanation. This condition ensures that $XP$ captures the essential features responsible for $x$’s outcome under the decision process $\Delta$. 

Algorithm~\ref{algo::CEalgorithm} computes a minimal abductive explanation for an individual $x$. The algorithm operates greedily: it attempts to remove each literal from $x$ and retains only those whose removal changes the outcome, thereby preserving minimal sufficiency. We provide a practical computation of an abductive explanation for an applicant in our case study (see Appendix~\ref{app::Compute_axp}).

\begin{algorithm}[ht]
    \caption{Compute-Explanation}
    \label{algo::CEalgorithm}
    \textbf{Input}: Decision process $\Delta$\\
    \textbf{Input}: Individual $x$ \\
    \textbf{Input}: Decision $d = \Delta(x)$ \\
    \textbf{Output}: $XP$, a minimal abductive explanation for $x$
    
    \begin{algorithmic}[1]
        \STATE $XP \leftarrow x$
        \FOR{each feature $v \in V$}
            \STATE $\ell \leftarrow x[v]$
            \STATE $XP \leftarrow XP \setminus \{\ell\}$
            %\IF{$\Delta(XP) \neq d$}
            \IF{$\exists x' \supseteq XP.\ \Delta(x') \neq d$}
                \STATE $XP \leftarrow XP \cup \{\ell\}$
            \ENDIF
        \ENDFOR
        \RETURN $XP$
    \end{algorithmic}
\end{algorithm}

\begin{theorem}\label{theo::algo1}
    The set of features returned by Algorithm~\ref{algo::CEalgorithm} 
    is a (subset-)minimal explanation for the individual.
\end{theorem}
We note that there can be multiple explanations for a single decision $\Delta(x)$
depending on the order in which the variables of the {\bf for} loop are chosen.  
The algorithm is not guaranteed, for instance, to return the minimal-cardinality explanations (with the smallest number of literals), 
nor is it something we are particularly interested in (minimal-subset explanations are sufficient).
\subsection{Bias in Decision Process}\label{section::bias}

Certain features, such as age, ethnicity, and gender, should not be used in decision-making as they constitute \emph{protected features}. We assume a scenario where there is only one protected feature, $p$. 
\begin{definition}\label{def::adnanBias}
    A decision $\Delta (x)$ is biased if it relies on the protected feature, which can be formally defined as follows:
\begin{displaymath}
    \exists x' \in  X_{(p)}^x.\quad 
    \Delta(x') \neq \Delta(x)
\end{displaymath}
\end{definition}

Definition~\ref{def::adnanBias} states that the decision of an individual is biased if their outcome would change if their protected feature were different. Consequently, the whole decision process is \emph{biased} if there exists at least one individual for whom the decision is biased.

Darwiche and Hirth~\cite{Adnan2020Reasons} established several theorems for detecting bias in decision processes using abductive explanations. 
Building on their formulation, a minimal explanation is considered \emph{biased} if it contains a protected feature. 
This foundation leads to the following theorems, which provide a formal basis for auditing bias in AI-driven decision processes.

\begin{theorem}
\label{biaseddecision}
A decision $\Delta(x)$ is \emph{biased} with respect to the individual $x$ if and only if all the minimal explanations that justify the decision of $x$ are \emph{biased}.
\end{theorem}

\begin{theorem}
\label{biaseddecisionprocess}
The decision function $\Delta$ is biased if and only if there exists an individual $x$ such that $\Delta(x)$ has a biased minimal explanation.
\end{theorem}

While these theorems offer a rigorous approach for detecting explicit bias using abductive explanations, they fall short in addressing more insidious forms of bias that arise when protected attributes exert influence indirectly through proxy variables. 
Such hidden pathways, leading to \emph{proxy discrimination}, require additional analysis, as discussed in the next section; all subsequent theorem proofs are deferred to Appendix~\ref{app:proofs}.

\section{Proxy Discrimination in Decision Process} \label{section::proxy discrimination}
In this section, we introduce the notion of implicit bias, known as \emph{proxy discrimination} within the framework of abductive explanations. 
To ensure clarity in our reasoning, we illustrate the discussion using a new applicant from our case study.
\begin{example}{\emph{Yahya}, a male applicant is proud that its decision is not biased but is it true?}\label{Yahya}

Consider the applicant \emph{Yahya}$= (A,G,\neg J, \neg H, \neg S, \neg B, \neg C, \neg D, P, M)$ being classified as good score risk by the decision process. He received the following explanation for its decision ($A,\neg S,\neg D, P, M$). 
Throughout this section, we examine whether \emph{Yahya}'s decision is biased.
\end{example}

\subsection{What is a Proxy Variable?}\label{subsection::proxy-variable}
Our definition of a proxy variable within the context of abductive explanations is 
grounded in the notion of background knowledge. 
This connection is motivated by the work of Liu et al.~\cite{yu2023}, 
who provide extensive insight into how background knowledge can guide 
the computation of realistic abductive explanations. 
In this paper, we define background knowledge as a set of constraints that enable 
the derivation of abductive explanations that are meaningful and applicable in real-world settings.

 \begin{definition}[Background Knowledge]
      \label{background-knowledge}
     A \emph{background knowledge} $K$ is a set of constraints satisfied by \emph{real individuals}. We denote the set of all real-world individuals, i.e., those that satisfy $K$, as $\bigl[ \bigl[ K \bigr] \bigr]$. The satisfaction of constraint $\phi$ by an individual $x$ is represented as $x \models \phi$.
     \begin{displaymath}
         \bigl[ \bigl[ K \bigr] \bigr]= \{ x \in X \mid x \models K \}.
     \end{displaymath}
 \end{definition}
 In our case study, the dataset reveals the following learned constraint as background knowledge: $k_1 = \neg( \neg G \land P \land M )$. $k_1$ captures that it is impossible to find a female applicant whose credit purpose is a car while also being single. We can then rewrite $k_1$ as \emph{if then} rule and obtain $k_1 = (P \land M \Rightarrow G)$.
% More generally, an explanation for the decision $\Delta(x)$ under background knowledge $K$ is a conjunction $XP$ satisfying:
%\begin{displaymath}
%    \forall x' \in [[K]].\quad XP \subseteq x' \Rightarrow \Delta(x) = \Delta(x').
%\end{displaymath}

Importantly, when background knowledge is considered, different decision processes may be equivalent for the set of real individuals. 
Let $\Delta$ and $\Delta'$ be two decision processes (e.g., a neural network and an SVM).
If they yield identical decisions for all individuals (not just real world ones), we say they are equal: $\forall x \in X, \Delta(x) = \Delta'(x)$, denoted as $\Delta = \Delta'$. 
Conversely, $\Delta \neq \Delta'$ means there exists some $x$ where $\Delta(x) \neq \Delta'(x)$.
Equivalence under background knowledge, written $\Delta \equiv_K \Delta'$, is defined as:
\begin{displaymath}
\forall x \in [[K]], \Delta(x) = \Delta'(x).
\end{displaymath}

In other words, $\Delta \equiv_K \Delta'$ means $\forall x \in X. \ \Delta(x) \neq \Delta'(x) \Rightarrow x \not\models K$. Similarly, we could rewrite the definition of an explanation under the background knowledge assumption as follows.
\begin{definition}[Abductive Explanation under BK]\label{explanation_under_BK}
$XP$ is an \emph{explanation} for decision $d$ under the background knowledge assumption if:\begin{displaymath}
    \forall x' \in [[K]].\quad XP \subseteq x' \Rightarrow \Delta(x') = d.
\end{displaymath}
\end{definition}

Returning to the concept of proxy variables, 
$k_1$ highlights a relationship between the attributes $P$, $M$, and the protected variable $G$. When $M=1 \Rightarrow G=1$. Thus we can say that $M=1$ is a context for proxy-discrimination. To formalize this observation, we introduce the concept of \emph{context} $\Phi$, which can be broadly defined as a set of conditions that must be satisfied for proxy discrimination to be considered.
%begin belona
\begin{definition}\label{def::variable-dependency}
    Any logical formula $\Phi$ is independent of a variable $v$ if
    \begin{displaymath}
         (\exists v.\Phi) \equiv \Phi.
    \end{displaymath}
\end{definition}
Independence means that the value of variable $v$ will never affect the evaluation of $\Phi$. Using Definition~\ref{def::variable-dependency}, we propose the following definition of a \emph{proxy variable}.
%end belona
\begin{definition}
     \label{proxy-variable}
     Variable $q$ is a \emph{proxy variable} of $p$ 
     iff there exists a propositional formula $\Phi$ (called a \emph{context})
     that does not mention $p$ or $q$
     in which knowing the value of $q$ can allow one to infer the value of $p$:
     \begin{displaymath}
       \exists \Phi, \nu_q, \nu_p.\quad 
       \left\{ 
         \begin{array}{ll}
           (i) & (\exists q.\ \Phi) \equiv \Phi\\
           (ii) & (\exists p.\ \Phi) \equiv \Phi\\
           (iii) &  K, \Phi  \not\models  \  (p = \nu_p)\\
           (iv) &  K, \Phi, (q=\nu_q) \  \models \  (p = \nu_p)\\
           (v) &  K, \Phi, (q=\nu_q) \  \not\models \  \bot.\\
        \end{array}
    \right.
     \end{displaymath}
\end{definition}
The first two conditions of Definition~\ref{proxy-variable} ensure that the context does not involve $q$ or $p$. This requirement prevents conclusions about the value of $p$ from being implicitly embedded within the context. The third condition states that $(p = \nu_p)$ cannot be derived from $K$ and $\Phi$ alone, while the fourth condition ensures that $(p = \nu_p)$ can be inferred when $(q = \nu_q)$ is added to the knowledge base. Finally, the last condition establishes the existence of at least one individual who satisfies these constraints, ensuring that the inference is not vacuous.

Now that we have established the definition of a proxy variable, we introduce two new theorems that explicitly address its influence within the decision process. These theorems provide formal criteria for detecting proxy discrimination in AI systems.

\begin{theorem}
    \label{existence-proxy discrimination}
    If the background knowledge $K$ does not mention the protected variable, there is no proxy variable.
\end{theorem}

\begin{theorem}\label{theo::equiv-bias-unbias}
  If the background knowledge $K$ allows for a proxy variable $q$ for the protected variable $p$, 
  then it is possible to construct two decision processes $\Delta$ and $\Delta'$ 
  that are equivalent ($\Delta \equiv_K \Delta'$) 
  such that $\Delta$ is biased according to the definition in subsection~\ref{section::bias}.
  and $\Delta'$ is not.
\end{theorem}

Theorem~\ref{theo::equiv-bias-unbias} demonstrates that when background knowledge is present, it is possible to find two equivalent decision processes, one biased and the other unbiased. 
This result leads to the conclusion that the conventional notion of bias from Definition~\ref{def::adnanBias} is inadequate
in the context of background knowledge.  
In the next subsection, we introduce a modified definition of bias, followed by a more comprehensive framework in the subsequent section.

\subsection{Proxy Discrimination in Decision Process}\label{subsection::proxy-discrimination}

At this point, we contend that the conventional notion of bias, as defined earlier, fails to capture cases involving proxy variables. 
Formally, an explanation is biased if it explicitly involves a protected attribute or, given background knowledge, a proxy variable that produces an equivalent discriminatory effect. 
We next formalise this extended notion through a new definition of bias in an explanation, followed by a corresponding definition of bias in a decision.

\begin{definition}[Factor of proxy discrimination]
    \label{proxy-explanation}
    An explanation $XP$ is a \emph{factor of proxy discrimination} 
    if it does not include the protected variable and 
    the real individuals it applies to share the same protected feature:
    \begin{displaymath}
        p \not\in vars(XP) \textnormal{ and }
        \exists \nu.\forall x \in \bigl[ \bigl[ K \bigr] \bigr].\ \
        K, XP \models p = \nu.
    \end{displaymath}
\end{definition}

\begin{definition}[Biased Decision under BK]
    \label{BK-bias}
    A decision process $\Delta$ is \emph{biased under background knowledge $K$} 
    if there exists an individual $x$ such that all its explanations are biased 
    either explicitly or through Definition~\ref{proxy-explanation}.
\end{definition}

This section demonstrates that a decision process may appear unbiased under the definition provided by Darwiche and Hirth~\cite{Adnan2020Reasons}, yet still exhibit bias through proxy discrimination.
This is illustrated by the case of applicant \emph{Yahya} in Example~\ref{Yahya}, whose decision is classified as \emph{background knowledge-aware bias} (\emph{BK-aware bias}).
Specifically, \emph{Yahya} is assessed as a good credit risk primarily due to his gender. 
Although the explanation does not explicitly mention gender, it applies only to male applicants. Thus, the decision process suffers from proxy discrimination, as explanations like in Example~\ref{Yahya} are valid exclusively for men.

\section{Unfairness Diagnosis in Decision Process} \label{section::unfairness}
\subsection{The Spaces Involved in the Decision Pipeline}
Our framework builds on Friedler et al.\cite{DBLP:journals/corr/FriedlerSV16} highlights that fairness depends on the interaction between different representational spaces in the decision pipeline, notably the \emph{feature space} and the \emph{observed space}.

The \emph{feature space} includes high-level, often abstract criteria relevant to the decision but not always directly measurable. The \emph{observed space}, in contrast, consists of measurable variables that approximate these criteria. Because features manifest differently across subgroups, individuals with the same protected attributes tend to cluster differently in the observed space.

\begin{figure}[!t]
\centering
\includegraphics[scale=0.7]{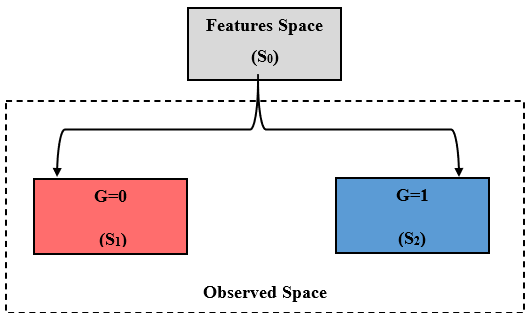}
\caption{Features and observed spaces of the running example}
\label{fig:spaces}
\end{figure}

Figure~\ref{fig:spaces} illustrates the interaction between the features and observed spaces using the running example.  Arrows in Figure~\ref{fig:spaces} represent how \emph{aptitudes} from the features space manifest in the observed space, leading to different properties depending on the subgroup to which the individual belongs. Notably, while the conditions in the features space are gender-neutral (because gender is not (should not be) an aptitude), the observed space is divided into subspaces based on gender, highlighting the importance of considering this distinction in decision-making.

Returning to abductive explanations, the concretization above yields explanations in the observed space. If an \emph{aptitude} is a sufficient condition for a decision, its equivalent in the observed space is, by Definition~\ref{def::AXP}, an abductive explanation. Fairness thus requires that individuals with the same \emph{aptitude} be treated equivalently in the observed space. The central question then becomes: how can this alignment be achieved? 
The next section addresses this question by presenting our proposed method.

\subsection{The Mapping Function and Aptitude Equivalence}

Our fairness framework operates under the assumption that individuals with equivalent aptitudes may be mapped differently depending on a protected variable (e.g., age, ethnicity, gender). Observable features result from a combination of \emph{fundamental} features and the protected variable, which we seek to neutralize.

\newcommand{\f}[0]{\textnormal{f}}

Formally, we assume a multi-set $S_\f$ of idealized aptitudes criteria that must be met to achieve a particular outcome. These aptitudes are assigned to two (or more) sets, $S_1$ and $S_2$, where assignment leads to abductive explanations $XP_1$ and $XP_2$, respectively, through a mapping function $M_{\f \to i}$. The concretization of an aptitude $xp_\f$ in $S_i$ results in an abductive explanation $XP_i$, which serves as an equivalent representation of the aptitude in that subgroup.

For simplicity, we assume that the mapping function $M_{\f \to i}$ is a bijection from $S_\f$ to $S_i$ (formally, $\forall xp_\f \in S_\f.\ \exists!\ XP_i \in S_i.\ XP_i = M_{\f \to i}(xp_\f)$). 

This implies that:
\begin{itemize}
    \item any a priori aptitude can be assigned to any group, and
    \item two different aptitudes from $S_\f$ will always be distinguishable in $S_i$ 
    ($xp_\f \neq xp'_\f \Rightarrow M_{\f \to i}(xp_\f) \neq M_{\f \to i}(xp'_\f)$).
\end{itemize}
One can guarantee this property by assuming, for instance, that there are infinitely many features in the space $S_i$, and that the decision process $\Delta$ ignores these extra features.

We then assume that decisions regarding individuals should be based solely on their aptitude.
% begin belona
\begin{definition}\label{defi::counterpart}
    Given a set of aptitudes $S_\f$ and two mapping functions $M_{\f \to 1}$ and $M_{\f \to 2}$, two abductive explanations $XP_1$ and $XP_2$ are equivalent noted $XP_1 \simeq XP_2$ if
    \begin{displaymath}
    \begin{array}{c}
        \exists xp_\f \in S_\f. \; \\
         M_{\f \to 1}(xp_\f) = XP_1 \land M_{\f \to 2}(xp_\f) = XP_2 
    \end{array}
    \end{displaymath}
%    \begin{align*}
%        &\forall xp_\f \in S_\f. \; \forall x_0, x_1 \\
%         &M_{\f \to 0}(xp_\f) \subseteq x_0 \;\land\;M_{\f \to 1}(xp_\f) \subseteq x_1 \Rightarrow\; \Delta(x_0) = \Delta(x_1)
%\end{align*}
\end{definition}
%end belona

\paragraph{Properties of the mapping function}
Because the mapping functions $M_{\f\to i}$ are assumed to be bijections, 
it is possible to inverse them.
%begin belona
In fact, we use the notation $M_{i\to\f}$ for the inverse function $M^{-1}_{\f\to i}$.
It is possible to rewrite Definition~\ref{defi::counterpart} as follows: 
\begin{displaymath}
  \begin{array}{c}
%    \forall x_1 \in S_1.\ \forall x_2 \in S_2.\\
    M_{1 \to \f}(XP_1) = M_{2 \to \f}(XP_2)\quad
   \Rightarrow\; XP_1 \simeq XP_2.
  \end{array}
\end{displaymath}
Actually, it is possible to chain the mapping functions
and write $M_{i\to j}$ for $M_{f\to j} \circ M_{ i\to f}$
which allows us to ignore $xp_\f$ altogether by stating the following.
\begin{displaymath}
  \begin{array}{c}
%    \forall x_1 \in S_1.\ \forall x_2 \in S_2.\\
    M_{1 \to 2}(XP_1) = XP_2\quad
    \Rightarrow\; XP_1 \simeq XP_2.
  \end{array}
\end{displaymath}
%end belona
This formulation has the benefit that the function $M_{i\to \f}$
which refers to the unobservable features
is no longer necessary to make equivalence between aptitudes. 

We link Definition~\ref{defi::counterpart} to the issue of proxy variables
through background knowledge.
Given a mapping function $M_{i\to \ell}$, where $\ell$ is either $j$ or $\f$,
the background knowledge of the population $S_i$ 
is a propositional formula $K$ 
such that $ \bigl[ \bigl[ K \bigr] \bigr] \subseteq \textnormal{dom}(M_{i \to \ell})$, 
i.e., $K$ is a formula that represents exactly all individuals $x$ 
for which $M_{i\to \ell}(XP_i)$ is defined. 

\begin{lemma}
\label{lemma:BK_domain_mapping}
  The background knowledge associated with the mapping function $M_{i \to j}$
  is the set of constraints $K$ such that $ \bigl[ \bigl[ K \bigr] \bigr]$ 
  evaluates to the union of the domains of these mapping functions.
\end{lemma}
\subsection{Defining and Auditing (Un)Fairness}

Fairness entails equivalent aptitudes receive similar treatment. The mapping function allows the equivalence between aptitudes from different subgroups. Therefore, we define fairness as follows: 
%Begin belona
\begin{definition}[Individual fairness] \label{fairness_decision}
Given two subgroups $S_1$ and $S_2$ and a mapping function $M_{1 \to 2}$, an individual $x_1 \in S_1$ is fairly treated by the decision process $\Delta$ if
\begin{displaymath}
\begin{array}{c}
    \exists XP_1 \subseteq x_1.\ \exists XP_2.\ XP_1 \simeq XP_2\\
    \forall x_2 \supseteq XP_2 \Rightarrow \Delta(x_1)=\Delta(x_2)
\end{array}
\end{displaymath}
    
\end{definition}
%end belona
Building on the previous definition, we now address fairness auditing at the individual level using the \emph{background knowledge} (BK). BK allows for \textit{a priori} verification of whether an individual is affected by proxy discrimination. We decompose the feature set $V$ into three parts: the base set $B$ (features independent of subgroup membership), the protected set $P$ (features defining subgroup membership), and the equivalence set $E$ (proxy features distinguishing subgroup members).

This decomposition helps identify decision explanations that support equivalence of aptitude across subgroups. We term such explanations \emph{criteria of (un)fairness}, formally defined as follows:

\begin{definition}[Criterion of (un)fairness]\label{fairness-criterion}
An explanation $XP$ is a \emph{criterion of (un)fairness} assessment if it allows the computation of a \emph{counterpart explanation} $XP'$.
\begin{displaymath}
    \exists XP'
    \left\{
   \begin{array}{l}
        XP'= M_{i \to j} (XP)\\
%        XP \cap    B \cup \{ \neg f | f \in B \} = XP' \cap    B \cup \{ \neg f | f \in B \} \\
        XP \cap    lits(B) = XP' \cap    lits(B) \\
   \end{array}
   \right.
\end{displaymath}
where $lits(B) = B \cup \{\neg v | v \in B\}$ is the set of literals defined over $B$.
\end{definition}
The definition~\ref{fairness-criterion} informs that two aptitudes are equivalents or counterparts if they share the same features of the base part ($B$).
\begin{figure}[!t]
\centering
\includegraphics[scale=0.5]{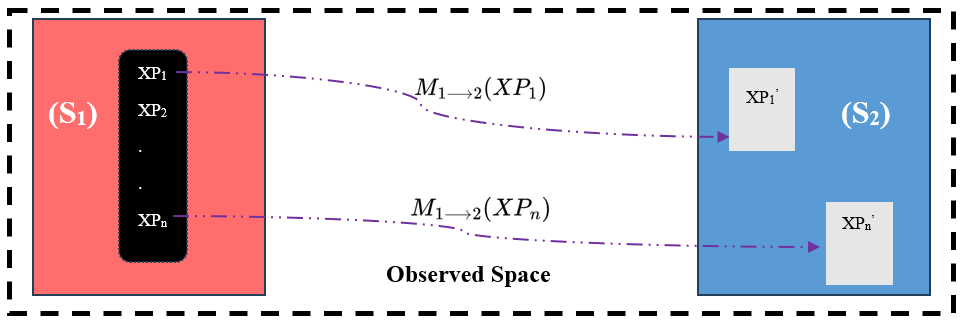}
\caption{Mapping function illustration}
\label{fig:new_mapping}
\end{figure}

The figure~\ref{fig:new_mapping} presents an individual $x$ from the subgroup ($S_1$) that has $ n$ explanations. $XP_1$ and $XP_n$ are criteria for fairness assessment as they allow computation of counterparts in the subgroup ($S_2$), whereas $XP_2$ is not.
\begin{theorem}\label{theo::individual_fairness}
    A decision $\Delta (x)$ of an individual ($x$) is fair if at least one of its explanation $XP$ is \emph{criterion  of (un)fairness} assessment and the associated counterpart explanation ($XP'$) yields to the same decision as $x$.
\end{theorem}

Within our framework, fairness is preserved since \emph{Yahya}’s explanation has a valid counterpart in the other subgroup, $XP' = (A, \neg G, \neg S, \neg D, \neg M)$, which leads to the same decision. Details of the procedure are provided in Appendix~\ref{app:yaya_example}. This illustrates the aptitude equivalence between an explanation associated with male applicants and a counterpart specific to female applicants.

Notably, this case underscores a critical insight: in certain situations, \emph{bias may be a necessary condition for fairness}. Specifically, to ensure subgroup equivalence, the abductive explanation used as a fairness criterion may need to include the protected attribute or its proxies.

\section{Discussion and Conclusion}\label{section::conclusion}
We proposed a formal framework based on abductive reasoning to assess fairness and proxy discrimination in AI decisions. Unlike statistical or counterfactual methods, our logic-based approach leverages domain knowledge to generate interpretable, case-specific explanations of bias. By formally characterizing proxy variables and fairness violations among individuals with equivalent aptitude, the framework enables the diagnosis of subtle, structural discrimination.

Our approach also supports intersectional fairness without requiring complex re-engineering: discrimination against subgroup $A \land B$ is flagged only if it is discriminatory for both, easing multi protected-attribute evaluation~\cite{liu2018fairmodmakingpredictive}. In addition, the framework can be extended to non-binary data by leveraging inflated formal abductive explanations, as proposed in the work of Izza et al.~\cite{Izza2023DeliveringIE}. While our examples draw on the German Credit dataset, future work will involve empirical studies on larger, real-world datasets to evaluate robustness and scalability. Furthermore, exploring interactive explanation tools~\cite{RevealNNInteractive2020} offers a promising direction to enhance transparency and accessibility for non-expert users.

\bibliographystyle{splncs04}
\bibliography{mybibliography}

\begin{thebibliography}{10}
\providecommand{\url}[1]{\texttt{#1}}
\providecommand{\urlprefix}{URL }
\providecommand{\doi}[1]{https://doi.org/#1}

\bibitem{barocas2016}
Barocas, S., Selbst, A.D.: Big data's disparate impact. California Law Review  \textbf{104}(3),  671--732 (2016)

\bibitem{Adnan2020Reasons}
Darwiche, A., Hirth, A.: On the reasons behind decisions. European Conference on Artificial Intelligence (ECAI)  (2020)

\bibitem{Emmert_Streib_2020}
Emmert-Streib, F., Yli-Harja, O., Dehmer, M.: Explainable artificial intelligence and machine learning: A reality rooted perspective. {WIREs} Data Mining and Knowledge Discovery  (2020)

\bibitem{EU2016}
{European Parliament and Council of the European Union}: Regulation (eu) 2016/679 of the {E}uropean {P}arliament and of the {C}ouncil of 27 {A}pril 2016 on the protection of natural persons with regard to the processing of personal data and on the free movement of such data, and repealing directive 95/46/ec. Official Journal of the European Union  (2016)

\bibitem{Ferrer2021}
Ferrer, X., Nuenen, T.v., Such, J.M., Coté, M., Criado, N.: Bias and discrimination in ai: A cross-disciplinary perspective. IEEE Technology and Society Magazine  \textbf{40}(2),  72--80 (2021)

\bibitem{DBLP:journals/corr/FriedlerSV16}
Friedler, S.A., Scheidegger, C., Venkatasubramanian, S.: The (im)possibility of fairness: Different value systems require different mechanisms for fair decision making. Commun. ACM  \textbf{64}(4),  136–143 (mar 2021)

\bibitem{gebru2018datasheets}
Gebru, T., Morgenstern, J., Vecchione, B., Vaughan, J.W., Wallach, H., Daumé~III, H., Crawford, K.: Datasheets for datasets. In: Proceedings of the FAccT. pp. 1--15. ACM (2018)

\bibitem{gupta2018proxy}
Gupta, S., Jung, C., Lykouris, T., Raghavan, M., Roth, A.: Proxy fairness. In: Proceedings of the 2018 ACM FAccT (FAccT). ACM (2018)

\bibitem{harman1965inference}
Harman, G.: The inference to the best explanation. The Philosophical Review  \textbf{74}(1),  88--95 (1965)

\bibitem{formalXAI2021}
Ignatiev, A., Narodytska, N., Asher, N., Marques-Silva, J.: From contrastive to abductive explanations and back again. In: Baldoni, M., Bandini, S. (eds.) XIXth International Conference of the Italian Association for Artificial Intelligence Virtual Event, November 25–27, 2020 Revised Selected Papers. pp. 335--355. Lecture Notes in Computer Science, Springer (2021)

\bibitem{Izza2023DeliveringIE}
Izza, Y., Ignatiev, A., Stuckey, P.J., Marques-Silva, J.: Delivering inflated explanations. In: Proceedings of the Thirty-Eighth AAAI. AAAI Press (2024)

\bibitem{johnson2025hardproxy}
Johnson, B.: The hard proxy problem: Proxies aren't intentional; they're intentional. Philosophical Studies  (2025)

\bibitem{keith2017abductive}
Josephson, J.R., Josephson, S.G. (eds.): Abductive Inference: Computation, Philosophy, Technology. Cambridge University Press, New York (1994)

\bibitem{lambers2020abductive}
Lambers, L., Sedlmair, M., Sorger, J., Streit, M., Aigner, W.: Abductive reasoning for explaining machine learning models. In: IEEE VIS Workshop on Explainable AI (XAI) (2020)

\bibitem{lindholm2024fair}
Lindholm, J., Manner, H., Stenberg, H.: What is fair? proxy discrimination vs. demographic disparities in insurance pricing. Scandinavian Actuarial Journal  \textbf{2024}(3),  213--239 (2024)

\bibitem{liu2018fairmodmakingpredictive}
Liu, J., Li, J., Liu, L., Le, T., Ye, F., Li, G.: Fairmod: making predictions fair in multiple protected attributes. Knowl. Inf. Syst.  \textbf{66}(3),  1861–1884 (Oct 2023)

\bibitem{Marques-Silva_Ignatiev_2022}
Marques-Silva, J., Ignatiev, A.: Delivering trustworthy ai through formal xai. Proceedings of the AAAI Conference on Artificial Intelligence  (2022)

\bibitem{mittelstadt2016ethics}
Mittelstadt, B.D., Allo, P., Taddeo, M., Wachter, S., Floridi, L.: The ethics of algorithms: Mapping the debate. Big Data \& Society  \textbf{3}(2),  2053951716679679 (2016)

\bibitem{RevealNNInteractive2020}
Myers, C.M., Freed, E., Pardo, L., Furqan, A., Risi, S., Zhu, J.: Revealing neural network bias to non-experts through interactive counterfactual examples (2020)

\bibitem{liu2025fairsense}
Raza, S., Chettiar, M.S., Yousefabadi, M., Khan, T., Lotif, M.: Fairsense-ai: Responsible ai meets sustainability (2025), \url{https://arxiv.org/abs/2503.02865}

\bibitem{shih2018symbolic}
Shih, A., Choi, A., Darwiche, A.: A symbolic approach to explaining bayesian network classifiers. In: Proceedings of the Twenty-Seventh IJCAI-18. pp. 5103--5111 (2018)

\bibitem{shih2019symbolic}
Shih, A., Choi, Y., Darwiche, A.: Symbolic logic meets machine learning: Interpretable explanations of black box models. In: Proceedings of the AAAI Conference on Artificial Intelligence. vol.~33, pp. 2931--2939 (2019)

\bibitem{tschantz2022proxy}
Tschantz, M.C.: What is proxy discrimination? In: Proceedings of the 2022 ACM Conference on FACCT. p. 1993–2003. FAccT '22, Association for Computing Machinery, New York, NY, USA (2022). \doi{10.1145/3531146.3533242}, \url{https://doi.org/10.1145/3531146.3533242}

\bibitem{VANDERWAA2021103404}
{van der Waa}, J., Nieuwburg, E., Cremers, A., Neerincx, M.: Evaluating xai: A comparison of rule-based and example-based explanations. Artificial Intelligence  (2021)

\bibitem{wachter2017counterfactual}
Wachter, S., Mittelstadt, B.D., Russell, C.: Counterfactual explanations without opening the black box: Automated decisions and the gdpr. Cybersecurity  (2017)

\bibitem{weerts2024unlawful}
Weerts, H., Kelly-Lyth, A., Binns, R., Adams-Prassl, J.: Unlawful proxy discrimination: A framework for challenging inherently discriminatory algorithms. In: Proceedings of the 2024 ACM FAccT. p. 1850–1860. FAccT '24, ACM (2024)

\bibitem{yu2023}
Yu, J., Ignatiev, A., Stuckey, P., Narodytska, N., Marques-Silva, J.: Eliminating the impossible, whatever remains must be true: On extracting and applying background knowledge in the context of formal explanations. Proceedings of the AAAI Conference on Artificial Intelligence  \textbf{37} (2023)

\end{thebibliography}
\appendix
\section{Case Study: A Decision Process for Classifying Individuals as Good or Bad Credit Risks}\label{example}

\subsection{The AI Model Used for the Decision-Making Process}\label{app:decision_process}
We illustrate our work using a simplified, binarized version of the German Credit Dataset, where numerical values above the median are set to 1 and others to 0, while categorical features take 1 for the most frequent value and 0 otherwise. The decision-making process is a single-layer neural network trained on this dataset, with the learned weights encoded into an SMT model to compute explanations.

\subsection{How to Compute a Formal Abductive Explanation}\label{app::Compute_axp}
 \begin{example} Let us compute a minimal abductive explanation for the decision of an individual \emph{Hawa}= $(A,\neg G,\neg J,\neg H,\neg S,B,\neg C,\neg D,P,M)$, who has been classified as a good credit risk ($\Delta(Hawa) = 1$) by the model.
\end{example}

We want to compute an explanation for \emph{Hawa}, represented by the vector $[1,0,0,0,0,1,0,0,1,1]$.
The first step (first row) of the table below verifies whether feature $A$ is necessary.
The check is positive (represented by $\top$ in column $\exists x'$);
thus $A$ is indeed necessary to reach the decision, therefore its initial value is reintroduced in the second row.
In the second step, $\neg G$ is shown to be unnecessary ($\bot$ in the last column)
and its value is therefore permanently removed.
Algorithm~\ref{algo::CEalgorithm} proceeds until all variables have been tested.
The final row contains the minimal explanation $A \land B \land \neg D \land P$.

% \begin{table}[!h]
\begin{center}
%  \centering
%  \caption{Computing an explanation for individual $[1,0,0,0,0,1,0,0,1,1]$}
%  \label{tab:table1}
  \begin{tabular}{l||l|l|l|l|l|l|l|l|l|l||l} 
      \textbf{Steps} & \textbf{A} & \textbf{G} & \textbf{J} & \textbf{H} & \textbf{S} & \textbf{B} & \textbf{C} & \textbf{D} & \textbf{P} & \textbf{M} & \textbf{$\exists x'$} \\
      \hline
      1 & ? & 0 & 0 & 0 & 0 & 1 & 0 & 0 & 1 & 1 & $\top$ \\
      2 & 1 & ? & 0 & 0 & 0 & 1 & 0 & 0 & 1 & 1 & $\bot$ \\
      3 & 1 & ? & ? & 0 & 0 & 1 & 0 & 0 & 1 & 1 & $\bot$ \\
      4 & 1 & ? & ? & ? & 0 & 1 & 0 & 0 & 1 & 1 & $\bot$ \\
      5 & 1 & ? & ? & ? & ? & 1 & 0 & 0 & 1 & 1 & $\bot$ \\
      6 & 1 & ? & ? & ? & ? & ? & 0 & 0 & 1 & 1 & $\top$ \\
      7 & 1 & ? & ? & ? & ? & 1 & ? & 0 & 1 & 1 & $\bot$ \\
      8 & 1 & ? & ? & ? & ? & 1 & ? & ? & 1 & 1 & $\top$ \\
      9 & 1 & ? & ? & ? & ? & 1 & ? & 0 & ? & 1 & $\top$ \\
      10 & 1 & ? & ? & ? & ? & 1 & ? & 0 & 1 & ? & $\bot$ \\
      final & 1 & ? & ? & ? & ? & 1 & ? & 0 & 1 & ? &  \\
    \end{tabular}
\end{center}
%\end{table}
\section{Theorems and Proofs}\label{app:proofs}
\begin{proof}
    Theorem~\ref{theo::algo1}: At any time during the algorithm, the current set of features $XP$ is an explanation.  Because we try to remove each feature from the set, and keep it only if it makes the set a non-explanation, we also have the guarantee that the resulting explanation is minimal.
\end{proof} 

\begin{proof}
Theorem~\ref{existence-proxy discrimination}:
Since none of the three premises in Def.~\ref{proxy-variable} mentions $p$, nothing can be deduced about the value of $p$.
\end{proof}

\begin{proof}
Theorem~\ref{theo::equiv-bias-unbias}:
Consider the context $\Phi$ and the values $\nu_q$ and $\nu_p$ that show that $q$ is a proxy for $p$ according to Def.~\ref{proxy-variable}.
From the last property of this definition, there exists an individual $x_0$
  who satisfies $K$, $\Phi$, such that $x_0[q] = \nu_q$, 
  and (from the second last property) such that $x_0[p] = \nu_p$.
  Now, define $x'_0$ as an individual identical to $x_0$ except for the value of $p$, which is inverted (i.e., $x'_0[v] = x_0[v]$ for all $v \in V$ except for $p$, where $x'_0[p] = \neg x_0[p]$). Since $x_0$ satisfies $\Phi$ and $\Phi$ does not involve $p$, it follows that $x'_0$ also satisfies $\Phi$. However, $x'_0$ does not satisfy $K$ because $(p = \nu_p)$ can be inferred from $K$, $\Phi$, and $(q = \nu_q)$.
  
  Now consider any unbiased decision process $\Delta$.
  Construct a new decision process $\Delta'$ as follows: 
  \begin{displaymath}
      \Delta'= \left\{
      \begin{array}{ll}
        \neg \Delta(x) & \textnormal{ if } x = x'_0\\
        \Delta(x) & \textnormal{ otherwise.}
      \end{array}
      \right.
  \end{displaymath}
  Clearly, $\Delta'$ exhibits bias, as it assigns different decisions to $x_0$ and $x'_0$, despite the fact that they differ only in their value of $p$. However, $\Delta$ and $\Delta'$ are equivalent under $K$ since $x'_0$ is not a real individual.
\end{proof}
\begin{proof}
  Lemma~\ref{lemma:BK_domain_mapping}:
  The background knowledge 
  is such that $\bigl[ \bigl[ K \bigr] \bigr]$ 
  represents real individuals.
  According to the mapping functions, an individual $x$ whose protected variable evaluates to $i$ and $XP_i$ as aptitude
  is associated with his counterpart that has the aptitude $M_{i\to j}(XP_i)$, 
  and $x$ is a real individual only if $M_{i \to j}(XP_i)$ is defined.  
  Therefore, real individuals that satisfy $XP_i$ are the domain of $M_{i \to j}$, 
  and the complete set of individuals is the union of these domains.
\end{proof}

\begin{proof}
Theorem~\ref{theo::individual_fairness}:
    assume $x$ is fairly treated (Def. ~\ref{fairness_decision}). Let us choose explanation $XP = x$ (which is an explanation for $x$).  $XP$ is a criterion of (un)fairness assessment since there exists a counterpart explanation of $XP$ (this counterpart explanation is $x'$ since $ M_{i \to j} (XP)$ $ =  M_{i \to j} (x) = x'$. Since $x$ is fairly treated, $\Delta(x') = \Delta(x)$ so $XP = x$ is one of the criterion mentioned by the theorem.
    
    On the other hand, if $x$ is unfairly treated, then any criterion $XP$ of $x$ is such that $M_{i \to j}(XP) = XP' \subseteq x'$.  Since $\Delta(x') \neq \Delta(x)$, $\Delta(XP') \ni \Delta(x')$ cannot equate $\Delta(XP) = \Delta(x')$.
    
\end{proof}

\section{Details of Unfairness Audit in Yahya Decision}\label{app:yaya_example}
\emph{Yahya}’s decision from section~\ref{section::proxy discrimination} is \emph{BK-aware biased}. His explanation was $XP = (A, \neg S, \neg D, P, M)$, with the feature set decomposed into base $B = \{A, J, H, S, C, D\}$, protected $P = \{G\}$, and equivalence $E = \{P, M\}$. To audit fairness, we construct $XP' = (A, \neg G, \neg S, \neg D, \neg M)$, representing a female applicant also classified as good score. Both explanations share the same base part  
\[
XP \cap (B \cup \{\neg f \mid f \in B\}) = XP' \cap (B \cup \{\neg f \mid f \in B\}) = (A \land \neg S \land \neg D),
\]  
which qualifies $XP$ as a fairness criterion. Since $XP'$ leads to the same favorable outcome, we conclude that \emph{Yahya}’s decision is fair. This shows the aptitude equivalence between male- and female-associated explanations.

\end{document}